\documentclass{INTERSPEECH2023}

\usepackage{microtype}
\usepackage{multirow}
\usepackage{multicol}
\usepackage{subfigure} 
\usepackage{CJKutf8}
\usepackage{booktabs}
\usepackage{CJK}
\usepackage{color}
\usepackage{enumitem}
\usepackage{amssymb}
\usepackage{array}

\usepackage{url}
\usepackage{arydshln}

\definecolor{forestgreen}{rgb}{0.13, 0.55, 0.13}
\definecolor{ao}{rgb}{0.0, 0.5, 0.0}
\definecolor{napiergreen}{rgb}{0.16, 0.5, 0.0}
\definecolor{brown}{rgb}{0.59, 0.29, 0.0}

\interspeechcameraready

\title{How Does Pretraining Improve Discourse-Aware Translation?}
\name{Zhihong Huang$^{1*}$~~~~~Longyue Wang$^{2*}$\thanks{* Longyue Wang and Zhihong Huang contributed equally to this work.}~~~~~Siyou Liu~~~~~Derek F. Wong$^1$}
\address{
  $^1$NLP$^2$CT Lab, University of Macau\\
  $^2$Tencent AI Lab}
\email{nlp2ct.zhihong@gmail.com, vinnylywang@tencent.com, helen.liu103@gmail.com, derekfw@um.edu.mo}

\begin{document}

\maketitle
 
\begin{abstract}
Pretrained language models (PLMs) have produced substantial improvements in discourse-aware neural machine translation (NMT), for example, improved coherence in spoken language translation.
However, the underlying reasons for their strong performance have not been well explained. 
To bridge this gap, we introduce a probing task to interpret the ability of PLMs to capture discourse relation knowledge. 
We validate three state-of-the-art PLMs across encoder-, decoder-, and encoder-decoder-based models. The analysis shows that (1) the ability of PLMs on discourse modelling varies from architecture and layer; (2) discourse elements in a text lead to different learning difficulties for PLMs.
Besides, we investigate the effects of different PLMs on spoken language translation. 
Through experiments on IWSLT2017 Chinese-English dataset, we empirically reveal that NMT models initialized from different layers of PLMs exhibit the same trends with the probing task. Our findings are instructive to understand how and when discourse knowledge in PLMs should work for downstream tasks.
\end{abstract}

\section{Introduction}

\label{sec:intro}

Translating spoken language is a significantly challenging task due to its inherent characteristics such as irregular expressions and discourse properties~\cite{wang2017exploiting,wang2019discourse,wang2023survey}. 
In recent years, discourse-aware neural machine translation (NMT) has performed better by initializing the Transformer-based \cite{vaswani2017attention} models with pretrained language models (PLMs) in encoder \cite{guo2020document}, decoder \cite{rothe2020leveraging}, both~\cite{wang-etal-2020-tencent} or themselves \cite{liu2020multilingual}.
The common assumption is that NMT models can utilize rich knowledge from PLMs to tackle complex discourse phenomena \cite{agrawal2018contextual}. 
For example, some works found that better-translated results often contain more connective words \cite{liu2020multilingual}, which can be classified as explicit with the non-tree-structure shallow discourse relations \cite{prasad2014reflections}. Table~\ref{tab:example} shows an example of discourse-aware translation. 

However, it is still unclear how discourse knowledge is embedded in PLMs, and when PLMs are leveraged in discourse-level NMT. 
Towards the understanding of PLMs, probing tasks are exploited to provide fine-grained analysis of model ability \cite{liu2019linguistic}. 
Related works either probed different PLMs about tree-structure discourse knowledge based on rhetorical structure theory (RST) \cite{carlson2003building,koto2021discourse}, or probed multilingual PLMs about part of the discourse relations \cite{kurfali2021probing}. Therefore, a more comprehensive study on the effects of discourse knowledge in PLMs on NMT is needed, particularly for spoken language translation.

\begin{CJK}{UTF8}{gbsn}
\begin{table}[t]
\centering
	\caption{An example of discourse-aware translation from IWSLT2017 dataset. The {\color{blue}token} is an {\em explicit} connective word while {\color{blue}$\copyright$} is {\em implicit} that is invisible to models. ``Inp.'' and ``Out.'' represent the Chinese input and English translation, respectively. As seen, a coherent translation should accurately transfer discourse relation between sentences from source to target language (e.g., {\color{red}words}).}
    \scalebox{0.94}{
    \begin{tabular}{c m{7cm}}
        \toprule
        \multirow{3}{*}{\bf Inp.} & 有一个志愿者在他呆的村庄里发明了这个小玩意儿\\
        [0.5ex]\cdashline{2-2}\noalign{\vskip 0.5ex}
        & {\color{blue}$\copyright$}就是把废纸 压缩成块状用来燃烧\\
        [0.5ex]\cdashline{2-2}\noalign{\vskip 0.5ex}
        & {\color{blue}但是}它烧得很慢\\
        \midrule
        \multirow{3}{*}{\bf Out.} & This is one such volunteer and this is a device that he had built in the village where he worked. \\
        [0.5ex]\cdashline{2-2}\noalign{\vskip 0.5ex}
        & {\bf \color{red}And} the idea was that you could take waste paper; you could compress it and make briquettes that could be used for fuel.\\
        [0.5ex]\cdashline{2-2}\noalign{\vskip 0.5ex}
        & {\bf \color{red}But} this device was very slow.\\
        \bottomrule
    \end{tabular}}
	\label{tab:example}
\end{table}
\end{CJK}

\begin{figure*}[t]
\centering
\includegraphics[width=0.8\linewidth]{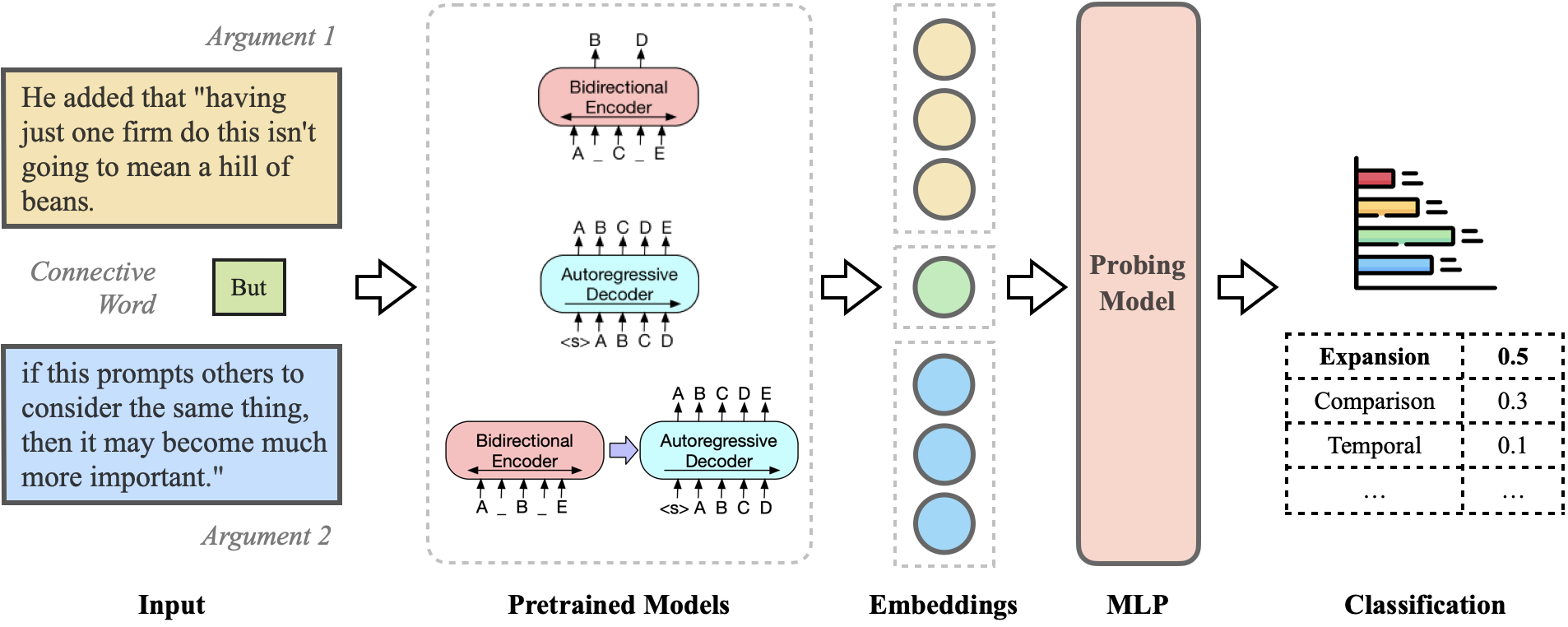}
\caption{The framework of our proposed probing task. The input is two adjacent sentences in a document, which are split into two spans (i.e., Argument 1 and 2) with explicit/implicit connective words. Second, we extract corresponding embeddings by feeding the input to a pretrained model. Third, the embeddings are used to train an MLP model for learning to classify the discourse relation between Argument 1 and 2. The classification accuracy is used to reflect the ability of PLM on modelling discourse.}
\label{fig:struc-probe}
\end{figure*}

To bridge this gap, we propose a method to probe the ability of advanced PLMs (i.e., BERT \cite{kenton2019bert}, BART \cite{lewis2020bart}, and GPT-2 \cite{radford2019language}) to capture discourse knowledge. 
Analysis results on PDTB dataset \cite{prasad2008penn} demonstrate that encoder-decoder-based PLMs perform best especially on higher layers (except for GPT-2). 
About the translation tasks, we leverage PLMs to discourse-aware NMT by utilizing part/all of their parameters. Experiment results on the Chinese-English IWSLT2017 dataset \cite{cettolo2017overview} show that (i) PLMs in the source language help more than that in the target language; (ii) NMT prefers PLMs with the same architecture (i.e., BART); (iii) NMT initialized with the single discourse-aware layer can achieve a close performance to using all layers; (iv) translation performance exhibits the same trends with the probing task at the layer level.

\section{Methodology}
\label{sec:format}

\subsection{Probing Discourse Knowledge}

Our probing tasks mainly focus on the shallow discourse relation in a sentence with two semantic arguments \cite{palmer2005proposition}, rather than consider the RST relations in several sentences. 
The shallow discourse relations can be characterized into five types: 
(1) {\em Explicit} relation means that the connective words in the sentence are visible;
(2) {\em Implicit} relation means that the sentence has no connective words but can be annotated manually;  
when the sentence has no connective word but shows a discourse relation by its expressions or entities, it contains (3) {\em AltLex} or (4) {\em EntRel} relation;
(5) {\em NoRel} relation means there is no discourse relation in the sentence. 
Further, {\em Explicit}, {\em Implicit}, and {\em AltLex} relations can be annotated with sense in three levels: {class}, {type}, and {subtype}. 
We list three examples below, whose words underlined are connective words, highlighted in italics are argument-1, and in boldface are argument-2.
\begin{itemize}
\small{
\item[(1)] \textit{It was a far safer deal for lenders} \underline{since} \textbf{NWA had a healthier cash flow and more collateral on hand.}
\item[(2)] \textit{Some have raised their cash positions to record levels}. \underline{(Implicit: BECAUSE)} \textbf{High cash positions help buffer a fund when the market falls.}
\item[(3)] \textit{Ms. Bartlett's previous work, which earned her an international reputation in the non-horticultural art world, often took gardens as its nominal subject.} \underline{(AltLex) \textbf{Mayhap this metaphorical connection made}} \textbf{the BPC Fine Arts Committee think she had a literal green thumb.}
}
\end{itemize}
In the sentence (2), the manual-annotation connective word \underline{BECAUSE} has an {\em Implicit} relation, a \textit{Contingency} class, a \textit{Cause} type, and a \textit{Reason} subtype. 
This kind of fine-grained sense is the discourse relation label of the sentences in our tasks dataset.

According to the shallow discourse relation, we propose two probing tasks to assess PLMs' ability to encode the discourse relations in different dimensional views:
(1) To interpret the overall ability of PLMs, we probe the whole sentence embeddings in the complete dataset, including the four discourse relations. 
In particular, we adopt two forms of BERT representation to extract the sentences: \textit{[CLS]} embeddings and average pooling embeddings of all tokens.
(2) To determine if the PLMs distribute the discourse knowledge in different linguistic elements, we individually probe the embeddings of connective words and sentiment arguments after obtaining the whole-sentence embeddings in the explicit data. 
The whole-sentence embeddings of implicit and AltLex data are also probed to compare with the explicit data.

The overview of the probing tasks' structure is shown in Figure~\ref{fig:struc-probe}.
First, 
we extract the representations of the whole sentences with the frozen-parameters PLMs.
Then we feed the different combinations of embeddings into two-layer MLPs (Multilayer Perceptron) to train probing models with the parameters updated.
The probing models are classifiers, and the labels are fine-grained sense relations of the input sentence.
All layers of three PLMs (i.e., BERT, GPT-2, and BART) are probed separately.
We evaluated the classification accuracy of the probing models.
Specifically, we treat BART‘s encoder layers as its first to sixth model layers, and decoder layers as its seventh to twelfth model layers.
The layer with a higher accuracy means that it is more capable of embedding the discourse knowledge.

\begin{table}[t]
\centering
\caption{Data statistics of the datasets on discourse probing and machine translation tasks. We calculate the sizes of training, validation, and testing sets in terms of sentence number. The number in brackets denotes the size of instances with explicit discourse relation.}
\begin{tabular}{lll}
\toprule
   \bf Task             & \textbf{Probing} & \textbf{Translation} \\
\midrule
{Dataset} & PDTB2.0 {\cite{prasad2008penn}}          & IWSLT2017   {\cite{cettolo2017overview}} \\
\midrule
Train   & 32,535 (18,459)  & 231,266      \\
Valid   & 1,436  (812)  & 879      \\
Test   & 1,928  (1,090)  & 6,046      \\
\bottomrule
\end{tabular}

\label{tab:dataset-table}
\end{table}

\begin{figure*}[t]
    \centering
    \subfigure[Overall Results of PLMs.]{
        \includegraphics[width=0.34\textwidth]{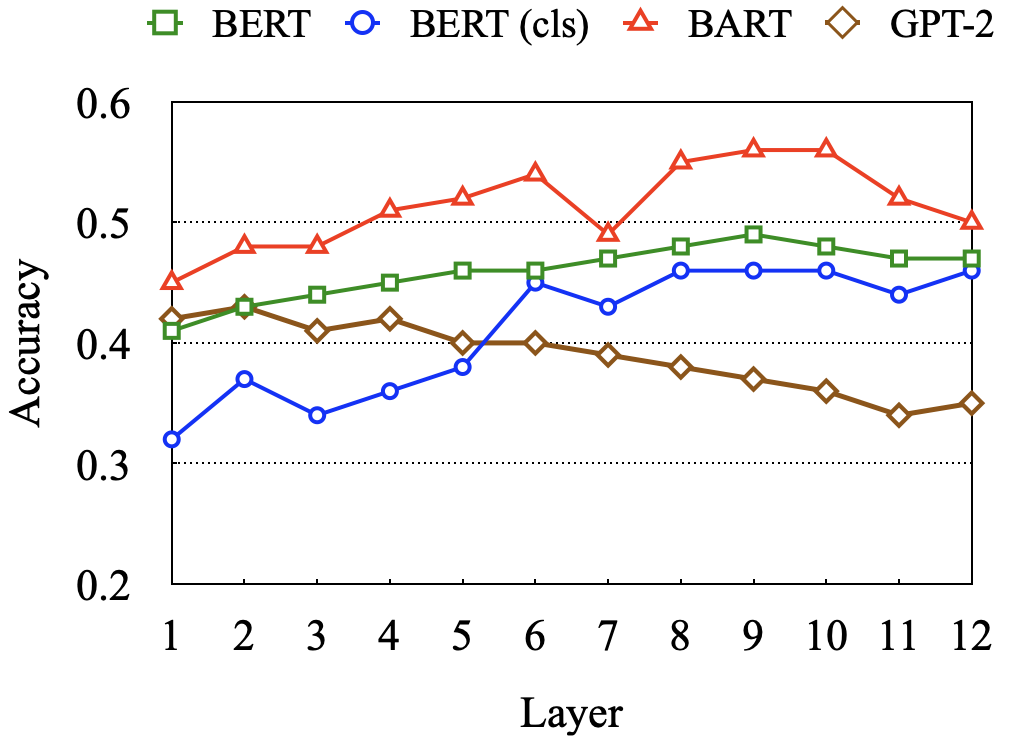}
        \label{picture-a}
    }
    \hspace{4em}
    \subfigure[Fine-grained Results of BART.]{
        \includegraphics[width=0.34\textwidth]{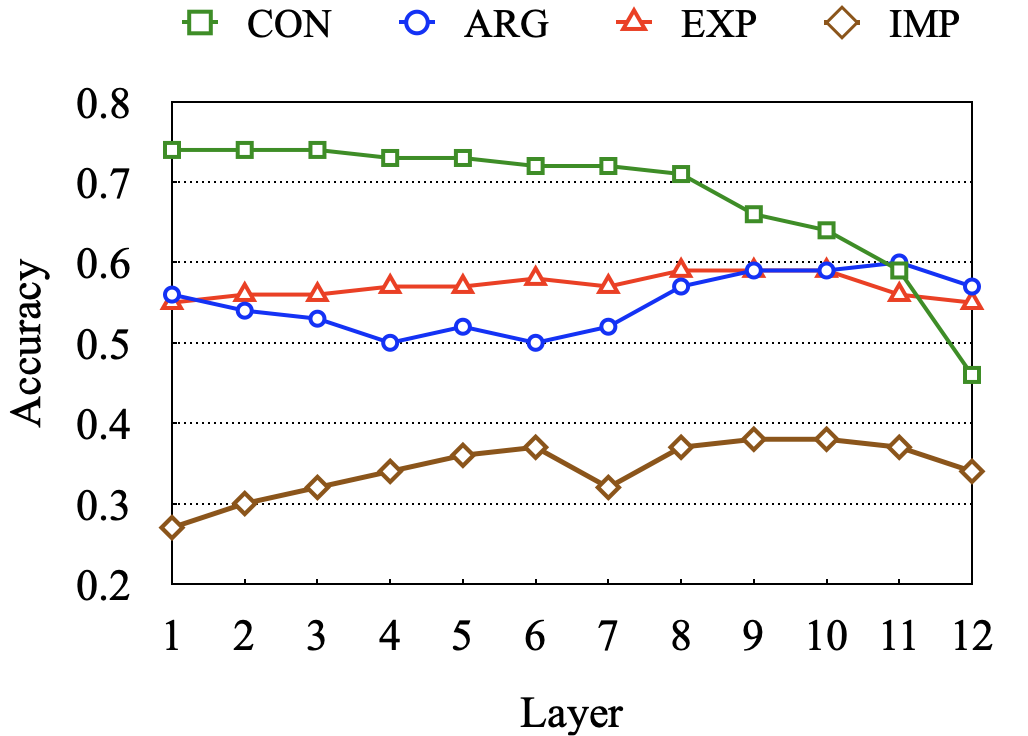}
        \label{picture-b}
    }
    \caption{Results of probing tasks for evaluating discourse properties embedded in different layers of representations in different pretrained models. 
    (a) shows the overall results of three PLMs on the complete dataset. 
    (b) shows the fine-grained results of BART by feeding parts of embeddings into the probing model. Note that, ``CON'' means only connective words while ``ARG'' indicates two arguments without connective words. We also report results on parts of data with explicit (``EXP''), and implicit  (``IMP'') discourse relations, respectively, including connective words.}
    \label{fig:fine-grained-prob}
\end{figure*}


\subsection{Discourse-Aware NMT with Pretraining}

Inspired by the work of Rothe et al.\cite{rothe2020leveraging}, we adopt the following strategies to leverage the PLMs to Transformer-based NMT models:
(1) For BERT models, we initialize the encoder with the PLMs and randomly initialize the decoder.
(2) For GPT-2 models, we initialize the decoder with the PLMs and randomly initialize the encoder.
(3) For BART models, we use them directly as a sequence-to-sequence model.
The three models are trained on Chinese data.
We also exploit Chinese and English versions for all three types of PLMs, and multilingual-BERT to investigate whether discourse knowledge in the source or target language is more significant to discourse-aware translation. 
For a fair comparison, all models have similar size of parameters.

We employ a document-level NMT model for the translation task, which can utilize context in the source language.
Following \cite{tiedemann2017neural}, we use one previous source sentence in the document as context information when translating each current sentence. 
Taking Table~\ref{tab:example} for instance, the first Chinese sentence can be used as context for translating the second one.
When the translation model can make better use of context information, that is, the translation of the current text contains more complete discourse knowledge, then its quality will be better.
We consider such a translation task to be a discourse-aware NMT.
Note that the sentence number of translation input is also consistent with that in the probing task, making the results of the two experiments comparable.
We use BLEU \cite{papineni2002bleu}, TER \cite{snover2006study}, and METEOR \cite{banerjee2005meteor} scores to measure the performance of the translation systems.

Besides, in a fine-grained view, we propose a method to investigate how PLMs perform in each layer when they are utilized for NMT.  
We train special discourse-aware NMT tasks by only updating the parameters of a specific layer each time, while the other layers' are frozen.
Based on the results of the probing tasks, we select the first, the middle, the last layer (i.e., layers 1, 6, 12), and the discourse-aware as the specific layer.
We wonder if the performance of NMT models with a single updated layer is related to the probing results. 
In this additional task, we only leverage the English-version PLMs, and keep the rest of the experiment settings the same as the normal discourse-aware translation task.

\section{Experiments}
\label{sec:pagestyle}

\subsection{Experimental Setup}
We summarize all data used in experiments in Table \ref{tab:dataset-table}.
For probing tasks, we conduct experiments on PDTB2.0 \cite{prasad2008penn}, which only contains English data. We simplify the discourse relation labels from 35 to 19 based on the strategy from \cite{xue2015conll}. 
\textit{EntRel} and \textit{NoRel} are also included as individual labels to consider as many shallow discourse phenomena as possible.

For discourse-aware NMT, we conduct experiments on IWSLT2017 Chinese-English dataset. IWSLT2017 is generated from TED talk, which is a spoken language dataset and has coherent sentences. 
According to our translation tasks, the adjacent sentences in the dataset can be combined as translation units for their discourse relation.
To make the model understand which sentence is context, a break token \textit{[SEP]} is inserted between two source sentences of every unit during the training. 
We use dev2010 for development and tst2010-2013 for testing from IWSLT dataset.
We train all NMT models up to 200K steps with 16 batch sizes.
Length penalty 
is set to 1.
Adam \cite{kingma2014adam} is used to optimize parameters with the 2e-5 learning rate.
The beam size of decoding is 4.

\begin{figure*}[t]
    \centering
    \subfigure[Results of BART.]{
        \includegraphics[width=0.32\textwidth]{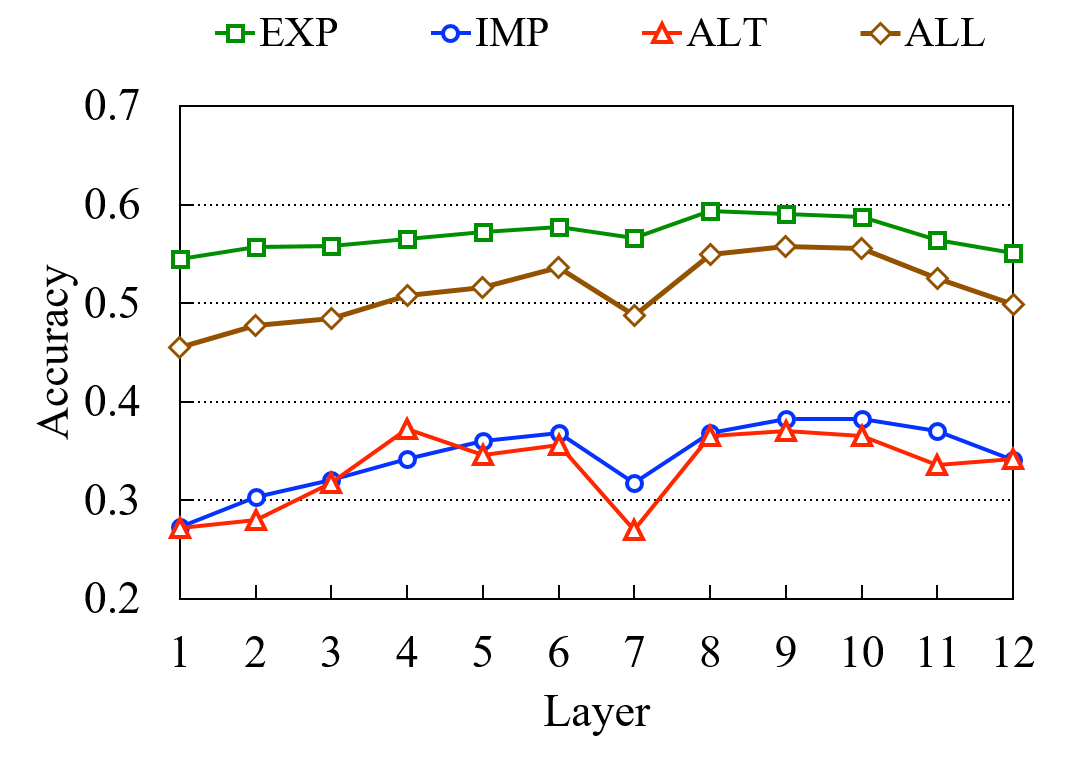}
        \label{picture-c}
    }
    \subfigure[Results of BERT.]{
        \includegraphics[width=0.32\textwidth]{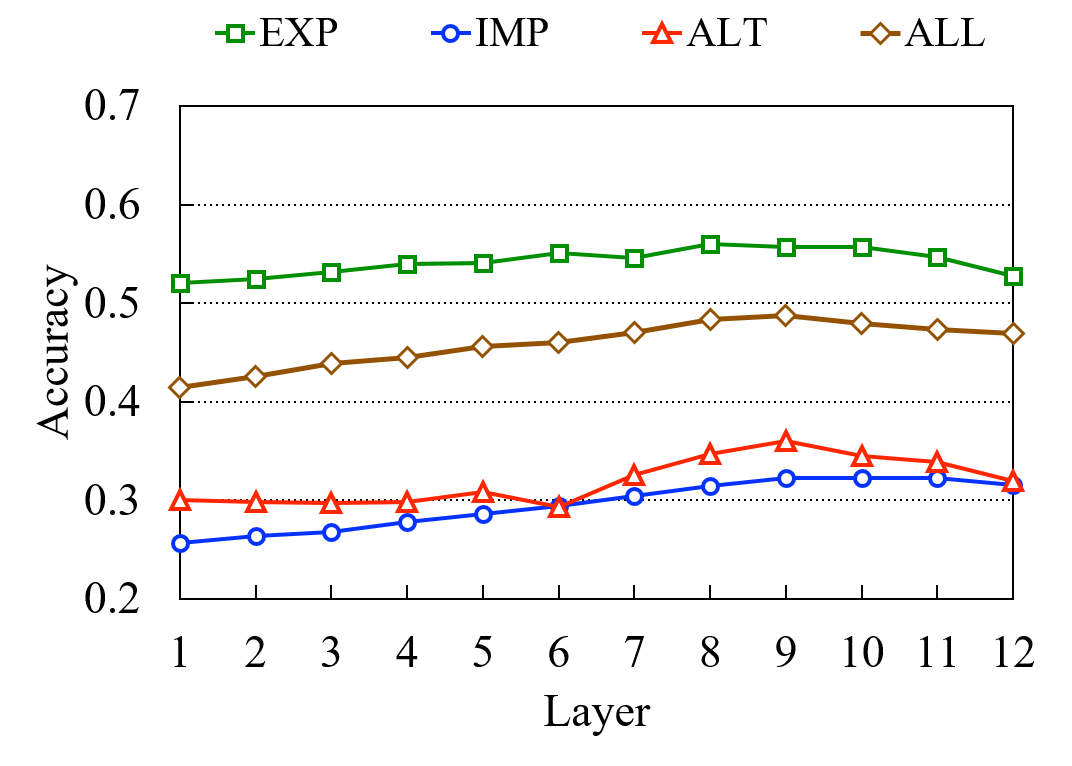}
        \label{picture-d}
    }
    \subfigure[Results of GPT2.]{
        \includegraphics[width=0.32\textwidth]{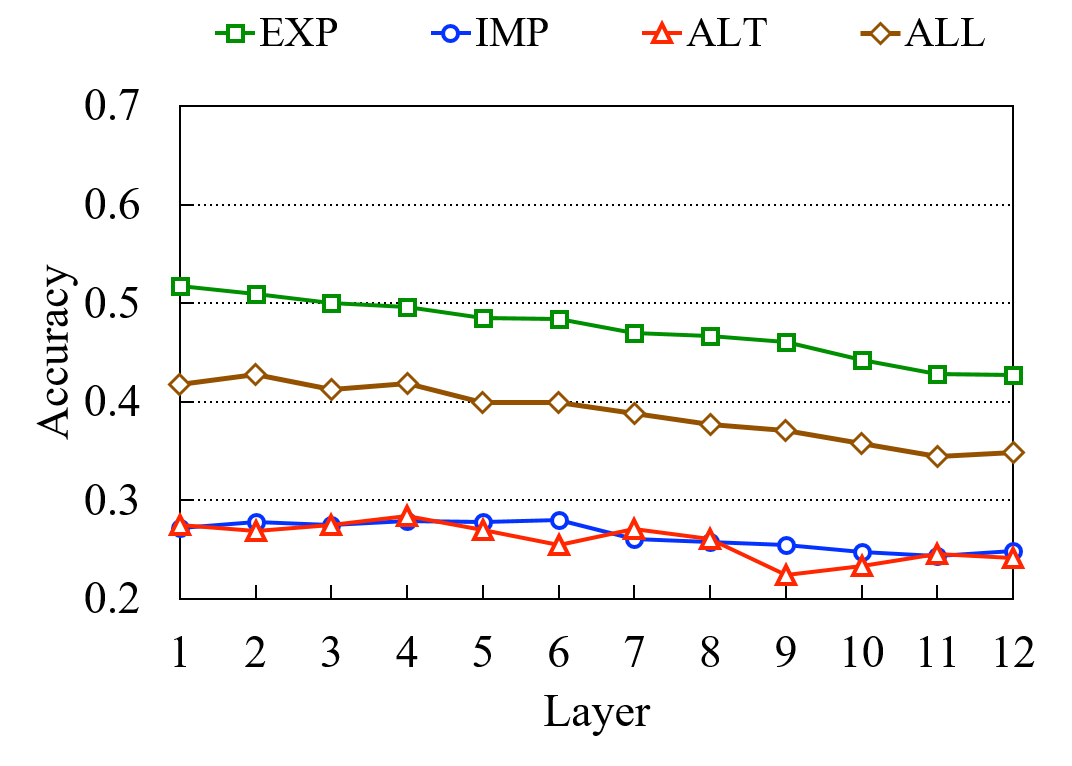}
        \label{picture-e}
    }
    \caption{Results of probing tasks for evaluating different discourse relations, including explicit (``EXP''), implicit  (``IMP''), and AltLex (``ALT'') discourse relation, embedded in different layers of representations in different pretrained models. We label the results of all data, including five discourse relations, as ``ALL'' }
    \label{fig:coarse-grained-prob}
\end{figure*} 

\subsection{Results of Probing Task}
\label{sec:probing-results}


In Figure~\ref{picture-a}, we present the first probing task performance for three basic models generated from the 12 layers. 
For the accuracy of all models' layers under 0.6, it seems that PLMs have a weaker capture ability in discourse than other linguistic knowledge \cite{vulic2020probing}.
It's clear that BART, as an encoder-decoder model, is the best among the three models. 
BART has a process of gradual increase in discourse knowledge in both all encoder layers and decoder layers 7-10.
In BERT, the average pooling of embeddings can extract more discourse knowledge than the \textit{[CLS]} embeddings.
Both BART and BERT have their discourse-aware layer in the ninth layer and have a significant decline after the discourse-aware layer.
Different from the other two PLMs, GPT-2 has its strongest discourse-aware capability in the first layer and then continually decreases in the after layers.

In Figure~\ref{picture-b}, we show the results of BART performed in the second probing task.
The curves of EXP and IMP both perform the same trend as BART (all data) in Figure~\ref{picture-a}.
This indicates that the relative capability between the different layers of PLM is the same in different kinds of discourse manifestations.
We can see that the accuracy of implicit data is much lower than that of the various embeddings of explicit data.
Without connective words, pretrained language models can not embed discourse knowledge well.
As the number of encoder layers increases, the discourse knowledge declines in both CON and ARG. But as the number of decoder layers increases, the discourse knowledge turns to rise in ARG, while that declines sharply in CON. 
CON contains larger discourse knowledge than EXP and ARG before layer 11 but is overtaken in layer 11 by ARG.

\subsection{Fine-grained Analysis}

As observed, the embeddings of connective words within most layers of PLMs encompass discourse knowledge.
In Figure~\ref{fig:coarse-grained-prob}, we further investigate the effects of three pretrained models on three types of discourse relations.
As seen, the accuracies of Implicit (IMP) and AltLex (ALT) relations are comparable and notably lower than that of the Explicit (EXP) relation.
Based on these consistent phenomena observed across the three models, we can reaffirm the conclusion drawn in Section~\ref{sec:probing-results} that connective words serve as essential elements for PLMs in comprehending discourse relations.
As demonstrated in \cite{liu2020multilingual}, mBART exhibits a preference for translating implicit discourse relations into explicit connective words, which in turn enhances the performance of document-level translation. This finding underscores the significant role of explicit discourse knowledge in PLMs in facilitating discourse-aware translation.


\subsection{Results of Translation Task}

\begin{table}[t]
\centering
\caption{Results of Chinese$\rightarrow$English translation task in terms of different evaluation metrics. We compare three pretrained models in different languages, including English (``EN''), Chinese (``ZH''), or multilingual (``Multi''). }
\begin{tabular}{cl rrr}
\toprule
\bf Lang. & \bf Model   & \textbf{BLEU}$^\uparrow$ & \textbf{TER}$^\downarrow$ & \textbf{METEOR}$^\uparrow$ \\ 
\midrule
\multirow{3}{*}{EN} & BERT         & 5.67       &   \textbf{83.52}    &     0.21       \\
 & BART         & \textbf{6.53}        &   83.96    &   \textbf{0.25}       \\
 & GPT-2         & 4.60        & 86.07       & 0.19          \\ 
 \midrule
\multirow{3}{*}{ZH} & BERT & 6.72         & 81.74       & 0.23          \\
&BART & \textbf{7.60}         & \textbf{78.57}       & \textbf{0.27}           \\
&GPT-2 & 4.64         & 86.18        & 0.19           \\ 
\midrule
Multi & BERT   & \textbf{12.90}       &  \textbf{75.82}     &  \textbf{0.32}             \\ 
\bottomrule
\end{tabular}
\label{tab:translation}
\end{table}

Table~\ref{tab:translation} shows the overall performance of PLMs leveraged in NMT models. 
Except for English BART's TER score which is weaker than that of English BERT, BART performs best on both language versions and all. 
GPT-2 is the worst.
As BART is an encoder-decoder model, we consider that document-level NMT task prefers PLMs with the same architecture. 
Besides, all three Chinese PLMs get better scores than the English PLMs. 
The multilingual-BERT even performs a huge improvement over all other PLMs. 
We consider that a PLM with the source language discourse properties will perform better than the target language.

Table~\ref{tab:ablation} shows the fine-grained results of the translation tasks. 
According to all three scores, the discourse-aware layer of PLMs revealed in probing tasks still performs better than the other layers in NMT tasks.
During the experiments, we found that only updating one layer's parameters can save around 13\% of training time, and the performance of the discourse-aware layer is close to that of all layers. 
The conclusions from Table~\ref{tab:translation} and Table~\ref{tab:ablation} show the same trend with the probing tasks.

\begin{table}[t]
\centering
\caption{Fine-grained results of translation task in Table~\ref{tab:translation}. We use only a specific layer of PLMs for initializing NMT models.}
\begin{tabular}{lccccc}
\toprule
\textbf{Layer} & \textbf{1} & \textbf{6} & \textbf{9} & \textbf{12} & \textbf{ALL}\\
\midrule
\multicolumn{6}{c}{\em BLEU$^\uparrow$} \\
BERT     & 4.86  & 4.89  & \textbf{5.04}  & 4.84   & \textbf{5.67}    \\
BART     & 5.27  & 5.46  & \textbf{6.03} & 5.66   & \textbf{6.53}\\
GPT-2     & \textbf{4.52}  & 3.69 & 3.51 &  3.37  & \textbf{4.60}     \\
\midrule
\multicolumn{6}{c}{\em TER$^\downarrow$} \\
BERT     & 85.12  & 84.97  & \textbf{84.92}  & \textbf{84.92}   & \textbf{83.52}    \\
BART     & 85.84  & 86.28  & \textbf{84.33} & 84.68   & \textbf{83.96}\\
GPT-2     & \textbf{85.55}  & 88.35 & 88.67 &  88.75  & \textbf{86.07}     \\
\midrule
\multicolumn{6}{c}{\em METEOR$^\uparrow$} \\
BERT     & 0.2002  & 0.2011  & \textbf{0.2019}  & 0.2011   & \textbf{0.2150}    \\
BART     & 0.2287  & 0.2215  & \textbf{0.2424} & 0.2374   & \textbf{0.2479}\\
GPT-2     &\textbf{0.1932}  & 0.1690  & 0.1676 &  0.1650  & \textbf{0.1916}     \\
\bottomrule
\end{tabular}
\label{tab:ablation}
\end{table}

\section{Conclusion and Future Work}

This paper analyzes discourse knowledge in pretrained language models for spoken language translation. 
The coincident trend in probing tasks and machine translation tasks reveals how the discourse knowledge changes in different layers of different pretrained language models when the model is leveraged in NMT.
Our work complements existing probing tasks about discourse knowledge in PLMs and provides a fine-grained interpretive perspective for the application of PLMs in document-level NMT.

In future work, we intend to delve deeper into the challenges of discourse-aware MT. Our plan is to leverage large language models (LLMs), which have shown great potential in dealing with complex tasks, to tackle this particular challenge \cite{wang2023document,lyu2023new}. Moreover, we plan to evaluate our approach on more discourse-aware tasks~\cite{wang2016naacl,wang2018learning,wang2019one,xu2022guofeng}.

\section{Acknowledgements}

This work was supported in part by the Science and Technology Development Fund, Macau SAR (Grant Nos. FDCT/060/2022/AFJ, FDCT/0070/2022/AMJ) and the Multi-year Research Grant from the University of Macau (Grant No. MYRG2020-00054-FST). This work was performed in part at SICC which is supported by SKL-IOTSC, and HPCC supported by ICTO of the University of Macau.






\bibliographystyle{IEEEtran}
\bibliography{INTERSPEECH2023}

\end{document}